\documentclass[lettersize,journal]{IEEEtran}
\usepackage{amsmath,amsfonts}
\usepackage{algorithmic}
\usepackage{algorithm}
\usepackage{array}
\usepackage[caption=false,font=normalsize,labelfont=sf,textfont=sf]{subfig}
\usepackage{textcomp}
\usepackage{stfloats}
\usepackage{url}
\usepackage{verbatim}
\usepackage{graphicx}
\usepackage{cite}
\usepackage{fontawesome5}
\usepackage{makecell}
\usepackage{multirow}
\usepackage{enumitem}
\usepackage{diagbox}
\usepackage{booktabs}
\usepackage{xcolor}
\usepackage[colorlinks,linkcolor=red,citecolor=green]{hyperref}
\usepackage{soul}

\begin{document}

\title{Joint Geometry–Appearance Human Reconstruction in a Unified Latent Space via Bridge Diffusion}

\author{Yingzhi Tang, Qijian Zhang, and Junhui Hou, \textit{Senior Member}, \textit{IEEE}
\thanks{This work was supported in part by the NSFC Excellent Young Scientists Fund 62422118, and in part by the Hong Kong Research Grants Council under Grants 11219324, 11202320, and 11219422. (\textit{Corresponding author: Junhui Hou})}
\thanks{Yingzhi Tang and Junhui Hou are with the Department of Computer Science, City University of Hong Kong, Hong Kong SAR. E-mail: yztang4-c@my.cityu.edu.hk; jh.hou@cityu.edu.hk.}
\thanks{Qijian Zhang is with Tencent Games, Shenzhen, Guangdong, China. Email: keegan.zqj@gmail.com.}
}



\maketitle

\begin{abstract}
	Achieving consistent and high-fidelity geometry and appearance reconstruction of 3D digital humans from a single RGB image is inherently a challenging task. Existing studies typically resort to decoupled pipelines for geometry estimation and appearance synthesis, often hindering unified reconstruction and causing inconsistencies. This paper introduces \textbf{JGA-LBD}, a novel framework that unifies the modeling of geometry and appearance into a joint latent representation and formulates the generation process as bridge diffusion. Observing that directly integrating heterogeneous input conditions (e.g., depth maps, SMPL models) leads to substantial training difficulties, we unify all conditions into the 3D Gaussian representations, which can be further compressed into a unified latent space through a shared sparse variational autoencoder (VAE). Subsequently, the specialized form of bridge diffusion enables to start with a partial observation of the target latent code and solely focuses on inferring the missing components. Finally, a dedicated decoding module extracts the complete 3D human geometric structure and renders novel views from the inferred latent representation. Experiments demonstrate that JGA-LBD outperforms current state-of-the-art approaches in terms of both geometry fidelity and appearance quality, including challenging in-the-wild scenarios. Our code will be made publicly available at \url{https://github.com/haiantyz/JGA-LBD}.
\end{abstract}


\section{Introduction}
\IEEEPARstart{R}{econstructing} high-fidelity digital humans from single-view RGB images is a fundamental problem in computer vision, with wide applications in virtual reality, gaming, autonomous driving, {\it etc.} Despite recent advances, achieving accurate reconstruction of both human geometry and appearance from a single image remains highly challenging, due to complex body shapes, diverse clothing, and severe self-occlusions.

Generally, there are two demands for the human reconstruction task, i.e., geometry reconstruction and appearance reconstruction.
Existing solutions for this task can be broadly grouped into implicit function-based \cite{saito2019pifu,saito2020pifuhd,zhang2023global,zhang2024sifu,ho2024sith}, explicit point-based \cite{tang2025human,2k2k,zhang2025idol}, and image-generation-based approaches \cite{zhang2025multigo, li2025pshuman}. Implicit function methods \cite{saito2019pifu,saito2020pifuhd,zhang2023global,zhang2024sifu,ho2024sith} extract pixel-aligned features, features from parametric human models such as SMPL, or other cues, and use MLPs to learn occupancy fields or SDFs for surface reconstruction. While effective for geometry reconstruction, they often fail to reconstruct accurate and fine-grained appearance because query points in 3D space rarely have exact color supervision; the closest surface point is typically used as a proxy, causing the model to learn an approximation rather than ground truth colors. Explicit point-based methods \cite{tang2025human,2k2k,zhang2025idol, tang2025hugdiffusion} represent humans with point clouds derived from RGB images, often via estimated depth maps. These approaches can reconstruct detailed geometry, but typically ignore appearance, or build appearance models on top of pre-reconstructed geometry or parametric models like SMPL models, resulting in multi-stage pipelines that may yields inconsistencies between geometry and appearance. Image-generation-based methods \cite{zhang2025multigo, li2025pshuman} leverage large generative models to synthesize multi-view images from a single input view, and then reconstruct geometry using techniques such as continuous remeshing \cite{palfinger2022continuous}. While promising, they also require multiple steps and are sensitive to artifacts in the synthesized views. Existing methods either struggle with appearance, focus solely on geometry or rely on complex, multi-stage pipelines. These limitations highlight two key requirements that remain unfulfilled: 
\begin{enumerate}
    \item high-quality ground-truth representations that simultaneously encode both geometry and appearance, while remaining amenable to effective network learning, are essential;
    \item a single-stage pipeline that jointly reconstructs geometry and appearance is required, as it naturally enforces consistency between geometric structure and appearance 
\end{enumerate}

For the first requirement, we adopt 3D Gaussians \cite{kerbl20233d} as the ground-truth representation, as they are explicit and can naturally capture both coarse geometry and fine-grained appearance. To further enable high-quality geometry and appearance encoding, we follow the two-stage ground-truth 3D Gaussian preparation strategy proposed in 
\cite{tang2025hugdiffusion}. By constraining the Gaussians to lie on the ground-truth surface, this strategy allows the representation to faithfully encode accurate geometry together with detailed appearance.

Therefore, we concentrate on the second requirement and reformulate the task as predicting a 3D Gaussian representation from a single RGB image in a single stage, enabling unified geometry reconstruction and novel view rendering.
One intuitive solution is to learn a compact latent representation of 3D Gaussians and conduct generative modeling in this latent space using diffusion models.
Nevertheless, existing approaches cannot be directly adopted: 3DShape2VecSet-based methods \cite{zhang20233dshape2vecset} are limited to implicit geometry and do not capture appearance, whereas Trellis \cite{xiang2025structured} depends on intermediate sparse structure generation, breaking the single-stage formulation.

In this work, we present JGA-LBD, a bridge diffusion model that learns in a unified latent space and enables single-step reconstruction of high-resolution 3D Gaussians of digital humans. Specifically, we design a sparse VAE jointly trained with geometry and appearance supervision, which maps input 3D Gaussians into compact latent representations. To fully exploit the rich information embedded in images, 
we extract two complementary modalities—depth estimation and SMPL prediction—from the input. However, their inherent discrepancies make direct utilization challenging. To address this, we introduce a modality unification module that transforms both modalities into 3D Gaussian representations, which are subsequently compressed into the same latent space by the sparse VAE. This design ensures that all subsequent diffusion learning is carried out in a unified latent space, substantially reducing training complexity. Building on this unified latent design, we reveal that bridge diffusion offers an unexpectedly suitable framework for human reconstruction, since the depth-conditioned latent naturally corresponds to a partial observation of the target latent code. Rather than generating from noise, the bridge diffusion model only needs to complete the missing components, thereby significantly reducing the generative difficulty and improving the quality of the learned latent representations. Finally, the decoded 3D Gaussians from the latent code enables both geometry surface extraction and high-quality novel-view rendering via splatting-based rasterization. Extensive experiments on two benchmarks, together with evaluations on in-the-wild images, consistently demonstrate that JGA-LBD outperforms state-of-the-art methods in both quantitative accuracy and qualitative visual realism.

In summary, the main contributions of this work are:
\begin{itemize}
    \item we design a sparse VAE that jointly compresses geometry and appearance of high-resolution 3D Gaussian representations into a compact latent code, overcoming prior methods that either focus solely on geometry or rely on additional sparse structural priors;
    \item  we introduce a modality unification module that converts depth estimation and SMPL prediction into latent structural guidance through a sparse U-Net and an SMPL inpainter, ensuring consistent conditioning across heterogeneous modalities; and 
    \item we adapt bridge latent diffusion to operate in the unified latent space,  enabling efficient single-stage generation of complete latent codes, simultaneously modeling geometry and appearance.
\end{itemize}

The remainder of this paper is organized as follows. Section \ref{sectionrelatedwork} provides a comprehensive review of the existing literature, including monocular depth estimation, point cloud generation, human pose and shape estimation and single-view human reconstruction. Section \ref{proposedmethod} introduces our proposed JGA-LBD in detail. In Section \ref{sectionexp}, we conduct extensive experiments and ablation studies to evaluate the effectiveness of JGA-LBD. Finally, Section \ref{sectionconclusion} concludes this paper.

\section{Related Work}
\label{sectionrelatedwork}
\subsection{Diffusion Models}
Diffusion models \cite{ho2020denoising} have achieved remarkable success in generative modeling across diverse domains, including image synthesis, video generation, and audio processing. The core principle is to learn data distributions by gradually denoising Gaussian noise. While powerful, directly performing the diffusion process in pixel space is computationally expensive and often redundant. To address this, latent diffusion \cite{rombach2022high} compresses the input into a compact latent space before applying the diffusion process, enabling efficient training while retaining high-quality generation. This paradigm has since become the standard for large-scale image and video diffusion models \cite{batifol2025flux,peebles2023scalable,melnik2024video}.

A parallel line of work focuses on conditional diffusion, which aims to guide generation with auxiliary inputs. Early approaches such as classifier guidance and classifier-free guidance \cite{dhariwal2021diffusion,ho2022classifier} inject conditional signals during the sampling process. Later methods, such as ControlNet \cite{zhang2023adding}, extend this idea by introducing trainable networks that modulate intermediate features with external conditions, achieving fine-grained controllability. Despite their effectiveness, these methods still initialize the diffusion process from Gaussian noise, which may limit their ability to fully exploit structured priors. Bridge diffusion models extend diffusion models by learning diffusion bridges that directly map between two endpoint distributions, rather than from pure noise to the target data distribution. DDBM \cite{zhou2023denoising} unifies score-based diffusion and OT-Flow-Matching, enabling diffusion models to handle non-noise inputs more naturally. BBDM \cite{li2023bbdm} reformulates image-to-image translation as a stochastic Brownian bridge process that directly models the transformation between two domains. Unlike conventional conditional diffusion models, BBDM \cite{li2023bbdm} performs bidirectional diffusion between source and target domains, alleviating domain gap issues.

Beyond diffusion-based generative models, flow-based methods \cite{lipman2022flow, batifol2025flux,liu2022flow} provide an alternative paradigm that learns deterministic and invertible mappings between data distributions with exact likelihood estimation. Flow Matching \cite{lipman2022flow} proposes a simulation-free approach for training CNFs by learning vector fields along fixed probability paths, unifying diffusion and flow-based models. By supporting non-diffusion paths such as optimal transport interpolation, it enables more efficient training and faster sampling with improved generalization. Rectified Flow \cite{liu2022flow} learns deterministic ODEs that transport between two distributions along nearly straight paths, offering an efficient alternative to diffusion-based generation. Its rectification procedure progressively improves transport efficiency, leading to fast and accurate inference with minimal discretization steps.

In this paper, we combine latent diffusion and bridge diffusion to address the human reconstruction problem for two main reasons. First, latent diffusion is employed to handle the large-scale 3D Gaussian representation, which typically contains over $100k$ primitives. Second, bridge diffusion enables the model to focus on generating occluded regions.

\subsection{3D Generative Models}
Generating 3D models is inherently more challenging than 2D image or video synthesis due to the diversity of 3D representations and the increased complexity of 3D supervision, giving rise to two main research directions: multi-view-based generation and direct 3D representation generation.

Multi-view approaches typically first synthesize multiple 2D views conditioned on the input using diffusion models, and then reconstruct 3D content from the generated multi-view images via techniques such as continuous remeshing \cite{palfinger2022continuous}, 3D Gaussian Splatting \cite{kerbl20233d}, or NeRF \cite{mildenhall2020nerf}. For example, Zero123 \cite{liu2023zero} employs Stable Diffusion to generate multi-view images, after which a NeRF is optimized—following the SJC formulation \cite{wang2023score}—to fit these synthesized views, and meshes are extracted via marching cubes from the learned density field. Leveraging the higher efficiency of 3D Gaussian Splatting (3DGS) compared to NeRF, methods such as DreamGaussian \cite{tang2023dreamgaussian} and LGM \cite{tang2024lgm} use image diffusion to produce multi-view images and subsequently fit 3DGS, however, these methods typically generate only a small number of 3D Gaussians (around $20k$ points), requiring additional techniques such as continuous remeshing to construct a mesh. Due to the limited point density, the synthesized novel-view images are often in low-resolution, resulting in reconstructed meshes with relatively coarse surfaces.

Direct 3D generation lacks a unified pipeline, as 3D information can be represented in various forms, such as point clouds, implicit representations, volumes, NeRF, or 3D Gaussian Splatting. Luo {\it et al.} \cite{luo2021diffusion} introduced a probabilistic model for point cloud generation by interpreting points as particles in a thermodynamic diffusion process.  DiffGS \cite{zhou2024diffgs} encodes a 3DGS scene into a triplane latent and learns in latent space with DiT \cite{peebles2023scalable}; CraftsMan3D \cite{li2024craftsman3d} compacts shapes into vecsets \cite{zhang20233dshape2vecset} and trains DiT to learn an implicit field before extracting meshes at inference; and Trellis \cite{xiang2025structured} compresses 3DGS with sparse CNNs to support multiple downstream representations but requires an additional stage to provide geometric cues (sparse structure) and cannot unify geometry and appearance within a single latent.

In contrast, our framework jointly compacts geometry and appearance into a unified latent representation and employs bridge diffusion to learn it in a single stage. 

\subsection{3D Human Reconstruction} 
\noindent \textbf{Implicit-based 3D Human Reconstruction.} PIFu \cite{saito2019pifu} is a pioneering work that reconstructs colored 3D humans using pixel-aligned features. Subsequent methods enhance implicit representations with additional cues: SiTH \cite{ho2024sith} generates a back-view image via ControlNet and uses a skinned mesh to resolve 3D ambiguity; GTA \cite{zhang2023global} introduces a ViT-based encoder–decoder to reconstruct clothed avatars with tri-plane features; and SIFU \cite{zhang2024sifu} leverages SMPL-X–guided cross-attention and a diffusion-based texture refinement pipeline to improve robustness in the wild. Despite these advances, implicit approaches lack ground-truth color supervision—appearance is approximated from the nearest surface point—limiting their ability to model high-fidelity textures.

\vspace{0.5em}
\noindent \textbf{Explicit-based 3D Human Reconstruction.} Natsume \cite{natsume2019siclope} proposed SiCloPe, which combines 2D silhouettes and 3D joints to generate consistent novel-view silhouettes for deep visual hull shape reconstruction, followed by back-view texture prediction using a conditional GAN. ECON \cite{xiu2022econ} combines implicit surface representation with explicit body regularization to reconstruct high-fidelity 3D humans in challenging poses and loose clothing. Han {\it et al.} \cite{2k2k} predicted global shape using a low-resolution depth network and local details via a part-wise image-to-normal network, which are then merged to produce high-resolution depth maps for full 3D reconstruction. Tang {\it et al.} \cite{tang2023high} reconstructed clothed humans from sparse RGB images using a volumetric coarse-to-fine strategy with 3D convolutions. HaP \cite{tang2025human} introduces a novel pipeline conditioned on depth maps and SMPL models, using diffusion models to generate human point clouds that are subsequently reconstructed into surfaces via screened Poisson \cite{kazhdan2013screened}.

\vspace{0.5em}
\noindent \textbf{3DGS-based 3D Human Reconstruction.}
Recently, 3DGS has emerged as a powerful explicit representation for human reconstruction. IDOL \cite{zhang2025idol} leverages a large-scale dataset and a transformer-based predictor to reconstruct animatable Gaussian avatars efficiently. MultiGo \cite{zhang2025multigo} introduces multi-level geometry learning with skeleton, joint, and wrinkle refinement, while LHM \cite{qiu2025lhm} employs a multimodal transformer to preserve fine clothing and facial details. \cite{chen2024generalizable} proposed a generate-then-refine pipeline and an HGM module to generate high-quality human 3D Gaussian attributes. HumanSplat \cite{pan2024humansplat} uses a video diffusion model for generating human 3D Gaussian attributes within a universal Transformer framework. HuGDiffusion \cite{tang2025hugdiffusion} introduces a 3D Gaussian attributes diffusion model, which is conditioned on SMPL semantic features and pixel-aligned features.  These methods demonstrate the strength of 3DGS in capturing both geometry and appearance, though challenges remain in compact representation learning and efficient generative modeling.

\vspace{0.5em}
\noindent \textbf{Optimization-based 3D Human Reconstruction.} TeCH \cite{huang2024tech}, Human-SGD \cite{albahar2023single}, WonderHuman \cite{wang2025wonderhuman} and GeneMAN \cite{wang2024geneman} can also tackle this task with good generalization ability. PSHuman \cite{li2025pshuman} and MagicMan \cite{he2025magicman} use diffusion models to generate multi-view images and reconstruct the 3D human with the generated multi-view images. Human-GIF \cite{hu2025humangif} formulates the human reconstruction task as a single-view conditioned human diffusion generation task, using the large diffusion models to generate missing information.

\section{Proposed Method}
\label{proposedmethod}

\begin{figure*}[t]
	\centering
	\includegraphics[width=6.9in]{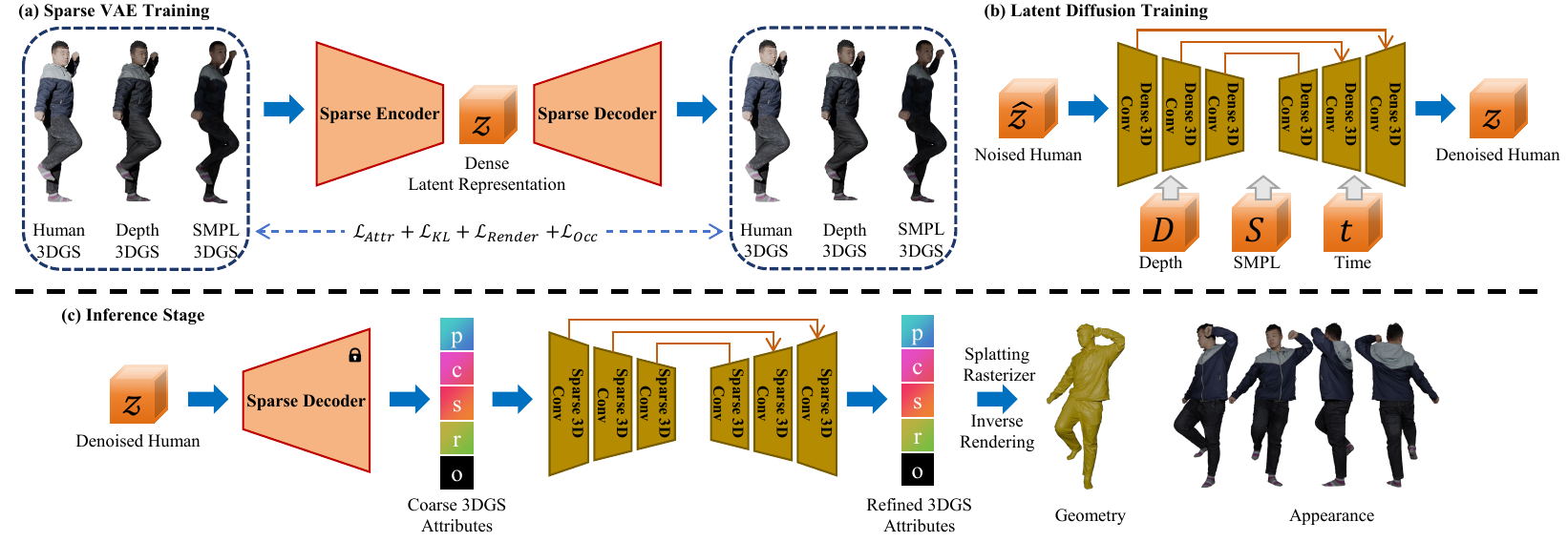}
	\caption{The pipeline of JGA-LBD. Given a single-view RGB image, depth and SMPL priors are converted into 3D Gaussians, which are compressed into latent codes by a sparse VAE. A bridge diffusion model generates latent codes conditioned on depth and SMPL priors, and the decoder refines them into a high-fidelity 3D Gaussian representation for surface reconstruction and novel-view rendering.  \textbf{Human 3DGS}: the ground-truth human 3D Gaussian attributes $\mathcal{G}$, \textbf{Depth 3DGS}: converted depth 3D Gaussian attributes $\mathcal{D}$, \textbf{SMPL 3DGS}: SMPL 3D Gaussian attributes $\mathcal{S}$.}
    \label{pipeline}
\end{figure*}

JGA-LBD is a 3DGS-based method that reconstructs a 3D Gaussian representation $\mathcal{G}=\{G_1,...,G_n\}$ from a single-view RGB image, where each element $G_i=\{p_i,c_i,s_i,r_i,o_i\}$ encodes its corresponding 3D Gaussian attributes such as position $p_i$, color $c_i$, scale $s_i$, rotation $r_i$ and opacity $o_i$. The resulting representation supports high-fidelity 3D surface reconstruction and novel-view appearance synthesis. 
As illustrated in Fig.~\ref{pipeline}, JGA-LBD consists of four key modules: 
\begin{enumerate}
    \item  a modality unification module to unify the depth maps and SMPL vertices into the same sparse 3D Gaussian format; 
    \item a sparse VAE that compresses both human 3D Gaussians and the converted conditions into latent codes in a unified latent space;
   \item  a bridge diffusion model that learns the distribution of latent human 3D Gaussians conditioned on structural priors; and 
    \item a decoder that transforms the denoised latent code back to a 3D Gaussian representation, followed by refinement to improve its fidelity before surface reconstruction and novel-view rendering. 
\end{enumerate}

Note that we follow the two-stage 3D Gaussian attribute preparation strategy proposed in HuGDiffusion \cite{tang2025hugdiffusion} to construct the ground-truth human 3D Gaussians for training. In this paper, we increase the number of 3D Gaussians from $20k$ to $200k$. All 3D Gaussians are then voxelized for sparse 3D CNN training, after voxelization, each scene will contain about $130k\sim150k$ Gaussians.

\subsection{Modality Unification Module}
Given a single RGB image, we infer two types of 3D information, namely a depth map and an SMPL model, which serve as conditional inputs for later diffusion model training. The depth map can be projected into a partial point cloud using camera parameters, while the SMPL model provides a complete geometric prior. Although both exist in 3D space, they belong to distinct modalities: the point cloud encodes 
$(x,y,z,r,g,b)$ values with appearance information, whereas the SMPL mesh contains only geometric vertices and faces. In HaP \cite{tang2025human}, sparse points are sampled from the depth-derived partial point cloud and the SMPL mesh using FPS, and no color information is utilized. As a result, these sparse geometric representations cannot be directly used as unified conditional inputs for joint geometry and appearance reconstruction. To address this issue, both modalities are first transformed into a consistent 3D Gaussian representation, enabling effective modality unification.

For the partial point cloud, we first perform nearest-neighbor search to associate each point with its closest Gaussian in the ground truth human 3D Gaussians (the same procedure is also applied when preparing the ground-truth SMPL 3D Gaussians), using these attributes as ground truth for supervision. A sparse U-Net based on Minkowski Engine \cite{choy20194d} is then trained to map $(r,g,b)$ values of the partial point cloud to 3D Gaussian attributes. For the SMPL mesh, we first project it onto the image plane, where only visible vertices receive color information, leaving occluded vertices uncolored. This partially observed mesh is then passed through another sparse Minkowski U-Net to predict complete 3D Gaussian attributes for all vertices, effectively generating 3D Gaussian representation. Note that the resulting SMPL 3D Gaussians are coarse and not intended as precise appearance supervision, this process primarily ensures that SMPL provides global structural guidance in a unified 3DGS format. By transforming heterogeneous modalities into 3D Gaussians, we obtain consistent and complementary conditional inputs for the diffusion process.

\subsection{Joint Geometry-Appearance Compression VAE}
Voxel is a common 3D representation that is compatible with standard CNNs. However, accurately representing a 3D object typically requires very high-resolution voxel grids (at least $512^3$), which is infeasible for training due to excessive GPU memory requirements. Inspired by latent diffusion \cite{rombach2022high}, we employ a sparse VAE to compress 3D Gaussians into a compact latent representation, enabling efficient modeling with standard CNNs in a unified latent space. 

Given the ground-truth human 3D Gaussians $\mathcal{G}$, converted depth 3D Gaussians $\mathcal{D}$, and SMPL 3D Gaussians $\mathcal{S}$, our goal is to encode them into a unified latent representation that jointly captures both geometry and appearance (all positions of Gaussians are voxelized before being fed into the sparse 3D CNN). Specifically, we build the sparse VAE with Minkowski Engine \cite{choy20194d}. The encoder $E(\cdot)$ of the sparse VAE consists of several ResNet blocks, the output $z$ of encoder $E(\cdot)$ serves as the ground truth for a diffusion model. To avoid learning a high-variance latent space, we impose a slight KL-penalty {$\mathcal{L}_{\text{KL}}$} to $z$ to make it learn a latent with a standard normal distribution.
The decoder $D(\cdot)$ is a key module in the sparse VAE, as it should decode the denoised $z$ of the diffusion model independently without any sparse structure cues like Trellis \cite{xiang2025structured}. Hence, we adopt the generative sparse transpose convolution layers to build the decoder $D(\cdot)$, which {enables generating} of new coordinates that do not need the cache coordinates from the encoder as in standard sparse transpose convolutions. It starts from $z$ and proceeds by progressively pruning excessive voxels with the occupancy loss {$\mathcal{L}_{\text{Occ}}$}:
\begin{equation}
\begin{aligned}
    \mathcal{L}_{\text{Occ}} = \;&  -\frac{1}{|\mathcal{V}|} \sum_{v \in \mathcal{V}} 
    \Big[
        o_g(v)\,\log o_p(v) 
        + \\ & \big(1 - o_g(v)\big)\,\log\big(1 - o_p(v)\big)
    \Big],
\end{aligned}
\end{equation}
where $\mathcal{V}$ denotes the set of activated voxels, 
$o_p(v) \in (0,1)$ represents the predicted occupancy probability, and 
$o_g(v) \in \{0,1\}$ is the ground-truth occupancy label;
and finally reaches the resolution of $\mathcal{G}$ after several layers. We use MSE loss to supervise the reconstruction of 3D Gaussian attributes, however, the predicted voxel grid and the ground-truth voxel grid are not strictly aligned and we cannot directly apply the MSE loss on the sparse tensors.  Converting sparse tensors into dense form introduces a vast number of non-active voxels (e.g., only about $130k\sim150k$ active voxels out of $512^3$), which seriously dilutes gradients and hinders effective learning. Therefore, we compute the MSE loss only on the intersection of active voxels between the prediction and the ground truth:
\begin{equation}
    \mathcal{L}_{\text{Attr}} = \frac{1}{|\mathcal{I}|} \sum_{i \in \mathcal{I}} \left\lVert \mathbf{a}_p(i) - \mathbf{a}_g(i) \right\rVert_2^2,
\end{equation}
where $\mathcal{I}$ denotes the intersection of active voxel indices, and $\mathbf{a}_p(i)$, $\mathbf{a}_g(i)$ are the predicted and gt 3D Gaussian attributes (i.e., $p,c,s,r,o$) at location $i$. However, supervising only on the intersection inevitably leaves certain regions unsupervised. To address this limitation, we introduce a loss by rendering the predicted 3DGS into 2D images and enforcing consistency with the ground-truth images. The rendering loss combines the L1 loss, SSIM loss and LPIPS loss:
\begin{equation}
\begin{aligned}
    \mathcal{L}_{\text{Render}} =  \lambda_1 \lVert \;& I_p - I_g \rVert_1 
+ \lambda_2 \big( 1 - \text{SSIM}(I_p, I_g) \big) \\
&+ \lambda_3 \, \text{LPIPS}(I_p, I_g),
\end{aligned}
\end{equation}
where $I_p$ and $I_g$ denote the rendered and gt images, respectively, and $\lambda_1,\lambda_2,\lambda_3$ are balancing weights. The overall training objective of the sparse VAE is:
\begin{equation}
    \mathcal{L}_{\text{VAE}} = \lambda_4 \mathcal{L}_{\text{KL}}+\lambda_5 \mathcal{L}_{\text{Occ}}+\lambda_6 \mathcal{L}_{\text{Attr}}+\lambda_7 \mathcal{L}_{\text{Render}}, 
\end{equation}
where $\lambda_4,\lambda_5,\lambda_6,\lambda_7$ are balancing weights. The encoded results of  $\mathcal{G}, \mathcal{D}$ and $\mathcal{S}$ are converted to dense latent representations for the diffusion training, denoted as $\mathcal{G}_L, \mathcal{D}_L$ and $\mathcal{S}_L$ respectively.

\subsection{Bridge Diffusion in Unified Latent Space}
Diffusion models are typically designed to transport data distributions into a standard Gaussian prior. However, in our setting, the depth-derived latent code $\mathcal{D}_L$ can be regarded as a structural subset of the full human Gaussian representation. Thus, instead of relying on naive diffusion models, we adopt the more powerful bridge diffusion model \cite{zhou2023denoising}, which learns a transport path between two arbitrary distributions. Specifically, the goal is to translate from the partial prior distribution $p_{\mathcal{D}_L}$ of the depth-derived latent code $\mathcal{D}_L$ to the target distribution $p_{\mathcal{G}_L}$, while being conditioned on the SMPL prior $\mathcal{S}_L$, the bridge diffusion process is encouraged to primarily model the missing information, thereby reducing the uncertainty and difficulty of the generation task.

Formally, a bridge diffusion process is represented by a sequence of time-indexed variables $\{x_t\}_{t=0}^T$. Using Doob’s $h$-transform \cite{doob1984classical}, the conditional stochastic bridge can be expressed as:
\begin{equation}
d x_t = f(x_t, t \mid \mathcal{S}_L)\,dt + g(t)^2\,h(x_t, t, y, T \mid \mathcal{S}_L)\,dt + g(t)\,d w_t,
\end{equation}
where {$f(x_t, t \mid \mathcal{S}_L)$ is the drift term and $g(t)$ is the diffusion coeff}, $x_0 \sim p_{\mathcal{G}_L}(x \mid \mathcal{S}_L)$, $x_T = y$, and $y \sim p_{\mathcal{D}_L}$. The term $h(x, t, y, T \mid \mathcal{S}_L) = \nabla_x \log p(x_T = y \mid x_t = x, \mathcal{S}_L)$ denotes the drift adjustment introduced by the $h$-transform to ensure that the process interpolates between the endpoints.
Reversing this bridge process yields the conditional reverse SDE:
\begin{equation} 
\begin{aligned}
d x_t = \;& \Big[ f(x_t, t \mid \mathcal{S}_L) - g(t)^2\big( U_\theta(x_t, t, y, T \mid \mathcal{S}_L) - \\ & h(x_t, t, y, T \mid \mathcal{S}_L) \big) \Big] dt + g(t)\, d\bar{w}_t,
\end{aligned}
\end{equation}
and the associated probability flow ODE:
\begin{equation}
\begin{aligned}
d x_t = \;& \Big[ f(x_t, t \mid \mathcal{S}_L) - g(t)^2\big( \tfrac{1}{2}U_\theta(x_t, t, y, T \mid \mathcal{S}_L) - \\ & h(x_t, t, y, T \mid \mathcal{S}_L) \big) \Big] dt,
\end{aligned}
\end{equation}
where $U_\theta(\cdot)$ denotes the neural network with parameters $\theta$ approximation of the bridge score function. To learn this score function, we adopt denoising bridge score matching, which minimizes the discrepancy between the predicted score and the closed-form conditional score of the Gaussian bridge:
\begin{equation}
\begin{aligned}
\mathcal{L}_{\theta}
= \;& \mathbb{E}_{x_t,x_0, x_T,t}
\Big[ w(t)\ \big\| U_\theta(x_t, x_T, t \mid \mathcal{S}_L) - \\ & \nabla_{x_t}\log q(x_t\mid x_0,x_T,\mathcal{S}_L) \big\|^2 \Big],
\end{aligned}
\end{equation}
where $w(t)$ denotes a time-dependent weighting function that adjusts the relative importance of different diffusion steps during training.

\textbf{Remark.} We augment each grid of the dense latent representation with an occupancy value $\{0,1\}$, allowing the bridge diffusion model to jointly learn both the latent features and their occupancy. During inference, the dense latent representation is converted into a sparse latent representation by retaining only the grids with predicted occupancy greater than 0.5.

\subsection{Decode Module}
We compress the original sparse 3D Gaussian attributes of size $(512^3, 20)$ into a latent representation of size $(64^3, 4)$ using a sparse VAE, significantly reducing GPU memory consumption and enabling feasible training. However, this aggressive compression inevitably leads to information loss, which is further exacerbated after the diffusion generation process. To alleviate this, following recent advances in large-scale 3D generative models \cite{xiang2025structured,ren2024xcube,li2024craftsman3d}, we append a Minkowski Engine--based U-Net after the VAE decoder to refine the outputs of the diffusion model.  
 
\begin{figure*}[t]
	\centering
	\includegraphics[width=6.9in]{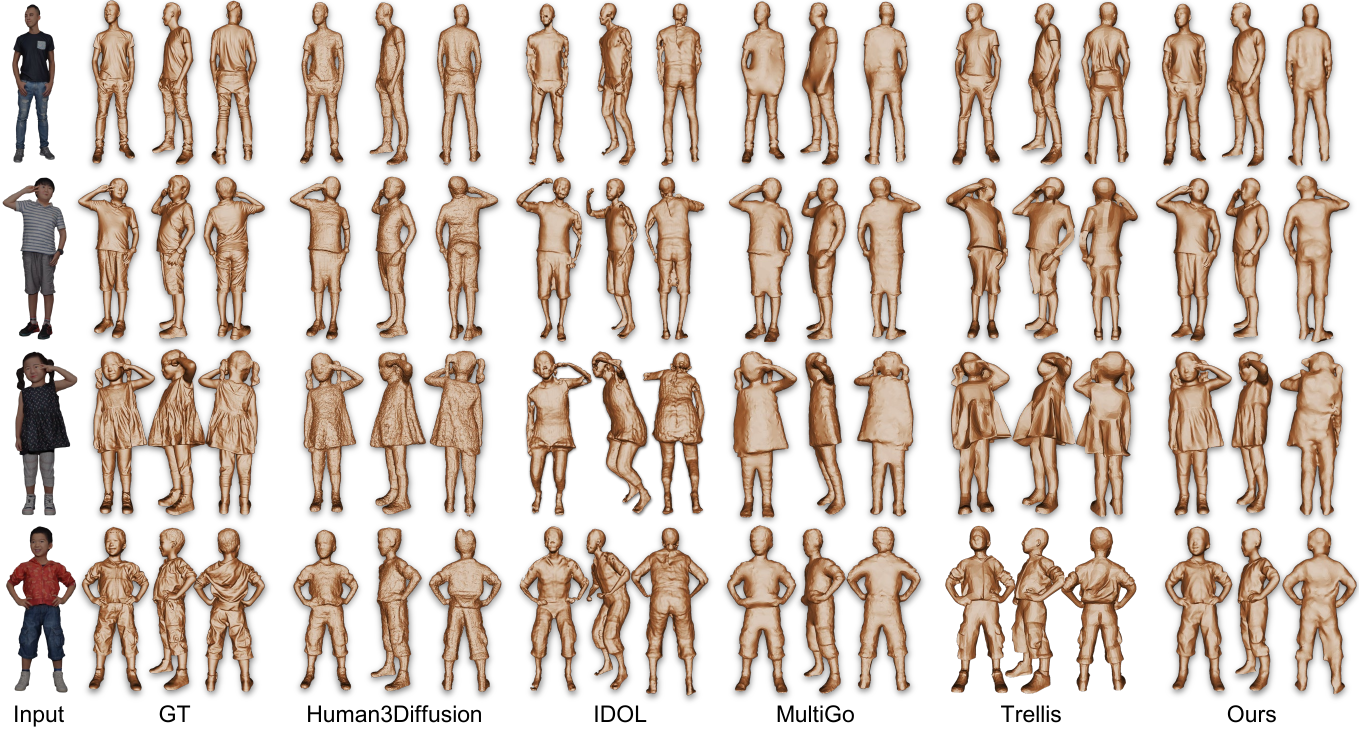}
	\caption{Geometry comparisons of our method against 3DGS-based methods, i.e., Human3Diffusion \cite{xue2024human}, IDOL \cite{zhang2025idol}, MultiGo \cite{zhang2025multigo} and Trellis \cite{xiang2025structured}. \textcolor{red}{\faSearch} Zoom in for details.  }
	\label{Geometrycompare}
\end{figure*}

\vspace{0.5em}
\noindent\textbf{Mesh Extraction.} For each reconstructed 3D Gaussians, we recover the human surface using its position attributes. Vertex normals are estimated via WNNC \cite{lin2024fast}, and the surface is reconstructed with screened Poisson \cite{kazhdan2013screened}. To further enhance geometric fidelity, the reconstructed surface is refined using depth supervision: the surface is rendered into a depth map with PyTorch3D, and an L1 loss is computed against the predicted depth map.

\vspace{0.5em}
\noindent\textbf{Novel View Synthesis.} For novel-view rendering, we adopt the standard 3D gaussian splatting pipeline, where the refined 3D Gaussians is rendered from arbitrary viewpoints using the corresponding camera parameters.  
\begin{table*}[t]
	\scriptsize
	\centering
	\renewcommand\arraystretch{1.25}
	\caption{Quantitative comparisons of different methods on 2K2K and CustomHuman. The best results are highlighted in \textbf{bold}. $\uparrow$: the higher the better. $\downarrow$: the lower the better.}
	\resizebox{\linewidth}{!}{\begin{tabular}{l|c|c|c|c|c|c|c|c|c|c|c|c} 
			\toprule
			\multicolumn{1}{c|}{\multirow{2}{*}{\diagbox{Method}{Metric}}} & \multicolumn{6}{c|}{2K2K} & \multicolumn{6}{c}{CustomHuman}    \\
			& \multicolumn{1}{c}{PSNR$\uparrow$} & \multicolumn{1}{c}{SSIM$\uparrow$} & \multicolumn{1}{c}{LPIPS$\downarrow$} & \multicolumn{1}{c}{CD$\downarrow$} & \multicolumn{1}{c}{P2S$\downarrow$} & \multicolumn{1}{c|}{Normal$\downarrow$} & \multicolumn{1}{c}{PSNR$\uparrow$} & \multicolumn{1}{c}{SSIM$\uparrow$} & \multicolumn{1}{c}{LPIPS$\downarrow$} & \multicolumn{1}{c}{CD$\downarrow$} & \multicolumn{1}{c}{P2S$\downarrow$} & \multicolumn{1}{c}{Normal$\downarrow$}\\
			\hline
			\hline
			GTA \textcolor{gray}{(NeurIPS 23)} & 24.15  & 0.921  & 0.080 & 1.156 & 1.114 & 2.127 & 28.86 & 0.920 & 0.088 & 1.249 & 1.123 & 2.552 \\ 
            SIFU \textcolor{gray}{(CVPR 24)}  & 23.47
			& 0.910 & 0.088 & 1.154 & 1.135 & 2.180 & 29.62 & 0.928 & 0.092 & 1.365 & 1.205 & 2.696 \\
			SiTH \textcolor{gray}{(CVPR 24)} & 24.30
			& 0.920 & 0.076 & 0.891 & 0.944 & 2.019 & 26.47 & 0.911 & 0.095 & 2.244 & 2.367 & 3.365 \\
            PSHuman \textcolor{gray}{(CVPR 25)} & 24.72 & 0.917 & 0.067 & 0.575 & 0.608 & 1.440 & 30.26 & 0.931 & 0.082 & 1.055&1.146 & 1.899\\
            \hline
             IDOL  \textcolor{gray}{(CVPR 25)} & 27.18 & 0.929 & 0.076& 1.095 & 1.138 & 2.454 & 31.02 & 0.934 & 0.076 & {1.119}& {1.188} & 2.416 \\ 
             Human3Diffusion \textcolor{gray}{(NeurIPS 24)} & 29.05 &0.942&0.062 & {0.503} & \textbf{0.415} & {1.429} &\textbf{33.75}&0.952&0.067 & {0.809}& {0.768} & {1.755} \\  
              MultiGO \textcolor{gray}{(CVPR 25)} &  {28.80} &  {0.939} &  {0.059}&  {0.636} &  {0.655} &  {1.474} &  {31.72} &  {0.934} & 0.075 & 1.750&1.809 & 2.440 \\  
              Trellis \textcolor{gray}{(CVPR 25)} & 25.47 &0.927&0.069 & 0.771 & 0.743 & 1.929 &31.33&0.934& {0.069} & 1.202& 1.219 &  {2.370} \\  
			\hline
			JGA-LBD & \textbf{30.16} & \textbf{0.946} & \textbf{0.055} & \textbf{0.489} & {0.507} & \textbf{1.202} & {33.44} & \textbf{0.957} & \textbf{0.061} & \textbf{0.674} & \textbf{0.670} & \textbf{1.469} \\ 
			\bottomrule
 	\end{tabular}}
	\label{sotaresults}
\end{table*}

\section{Experiments}
\label{sectionexp}
\subsection{Datasets and Evaluation Metrics}
\noindent\textbf{Datasets.} We conducted experiments on Thuman2.1 \cite{yu2021function4d}, 2K2K \cite{2k2k} and CustomHuman \cite{ho2023learning}. Specifically, 1600 scans from Thuman2.1 were used as ground truth to prepare the ground truth 3D Gaussian attributes, which were used for training the sparse VAE and the bridge latent diffusion model. For evaluation, we used 25 scans from 2K2K and 40 scans from CustomHuman. In addition, we evaluated the generalization ability of JGA-LBD on in-the-wild images collected from the Internet.

\vspace{0.5em}
\noindent\textbf{Evaluation metrics. } All 3DGS and mesh outputs are normalized to the cube $(-1,1)$. For appearance reconstruction, we report peak signal-to-noise ratio (PSNR), structural similarity index (SSIM), and learned perceptual image patch similarity (LPIPS). For geometry reconstruction, we evaluate Chamfer distance (CD), point-to-surface distance (P2S), and normal error.\\

\begin{figure*}[t]
	\centering
	\includegraphics[width=6.9in]{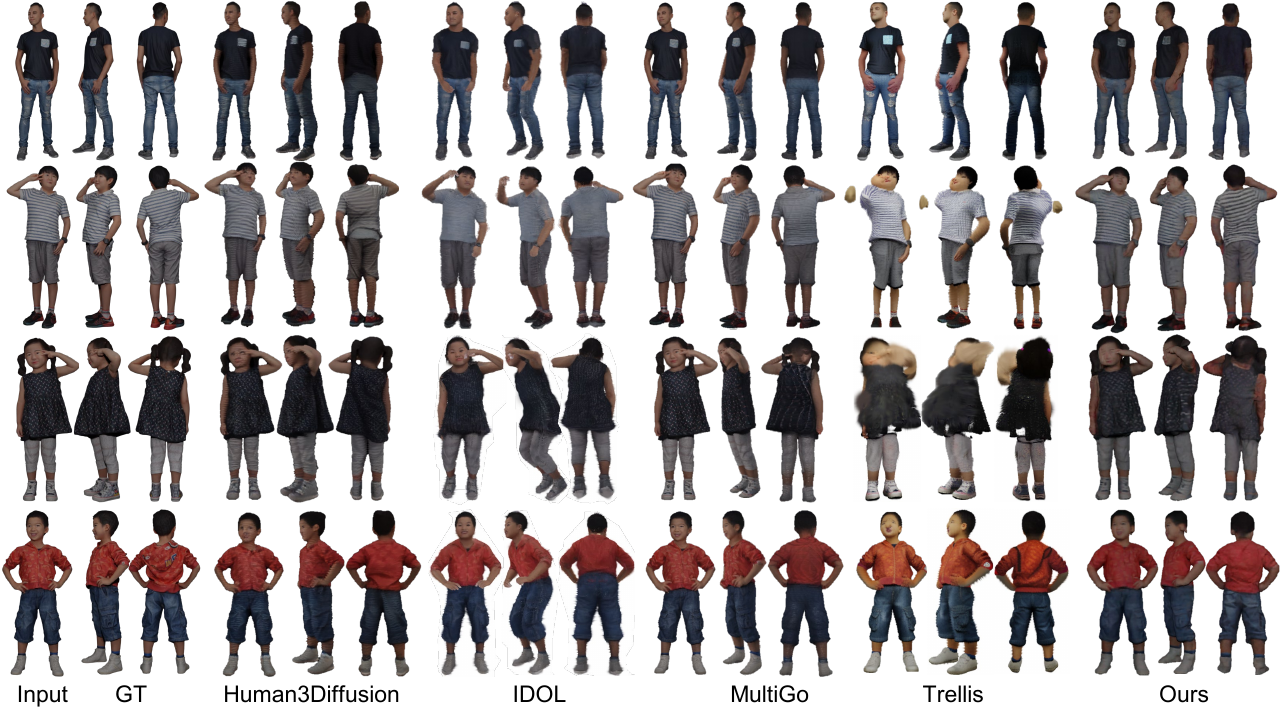}
	\caption{Appearance comparisons of our method against 3DGS-based methods, {\it i.e.}, Human3Diffusion \cite{xue2024human}, IDOL \cite{zhang2025idol}, MultiGo \cite{zhang2025multigo} and Trellis \cite{xiang2025structured}. \textcolor{red}{\faSearch} Zoom in for details.}
	\label{visualcompare}
\end{figure*}

\begin{figure*}[t]
	\centering
	\includegraphics[width=6in]{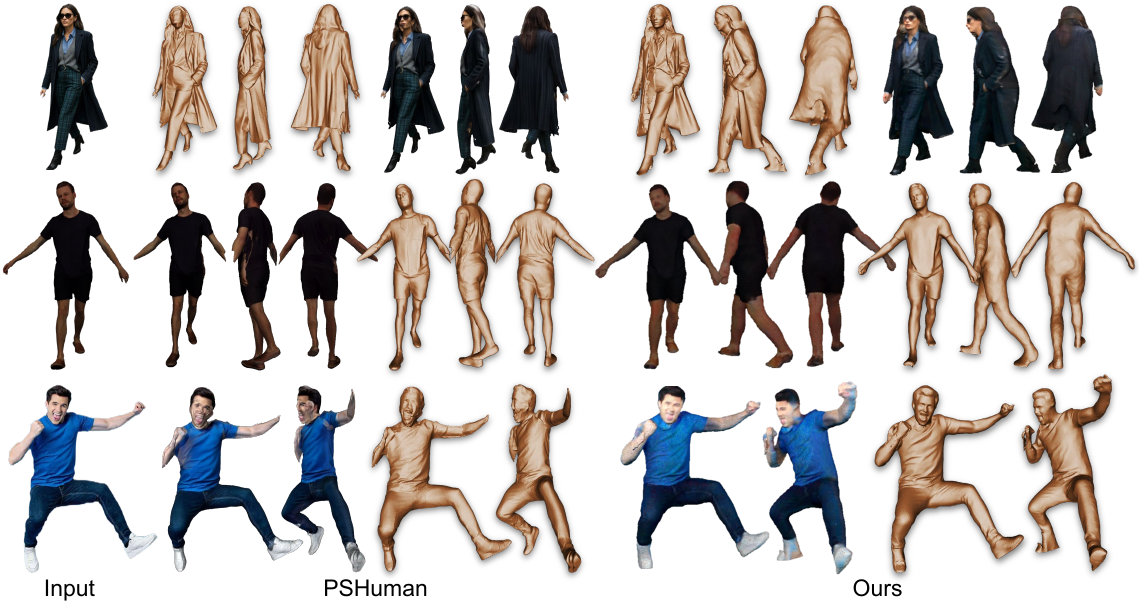}
	\caption{Geometry and appearance comparisons of our method against PSHuman \cite{li2025pshuman}. \textcolor{red}{\faSearch} Zoom in for details.}
	\label{pshumancompare}
\end{figure*}

\begin{figure*}[t]
	\centering
	\includegraphics[width=6.9in]{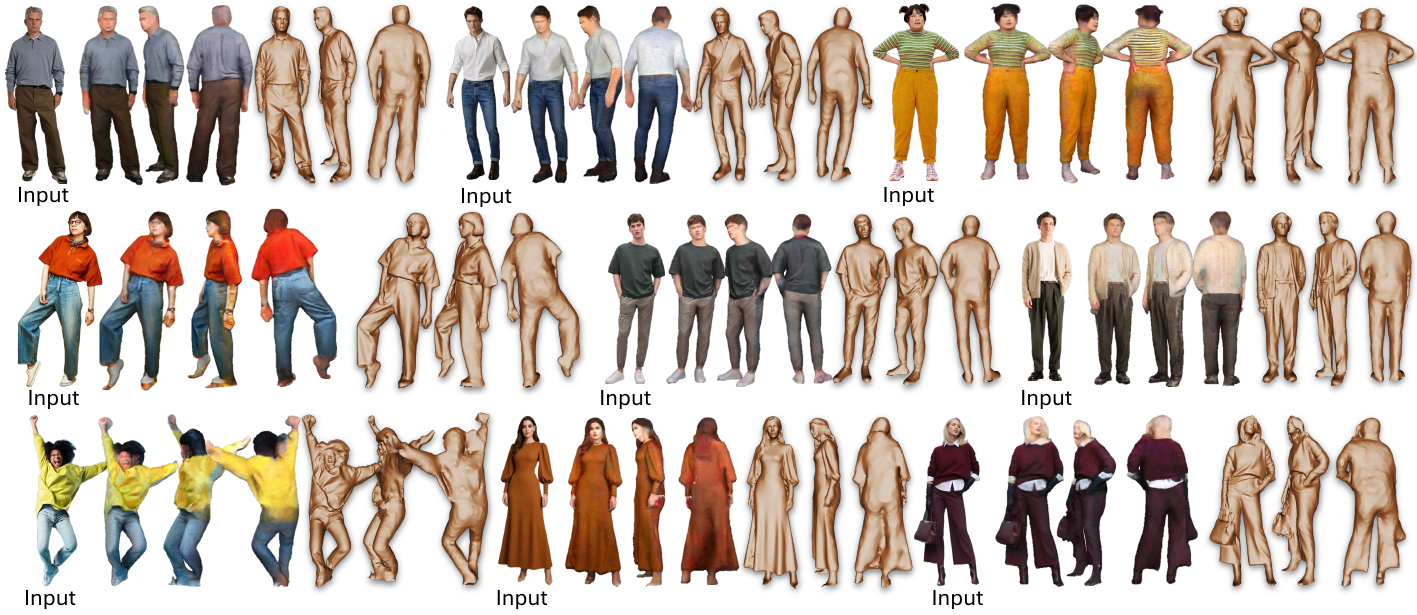} 
	\caption{The reconstructed results of our JGA-LBD on in-the-wild images. \textcolor{red}{\faSearch} Zoom in for details.}
	\label{inthewild}
\end{figure*}

\begin{figure*}[t]
	\centering
	\includegraphics[width=7in]{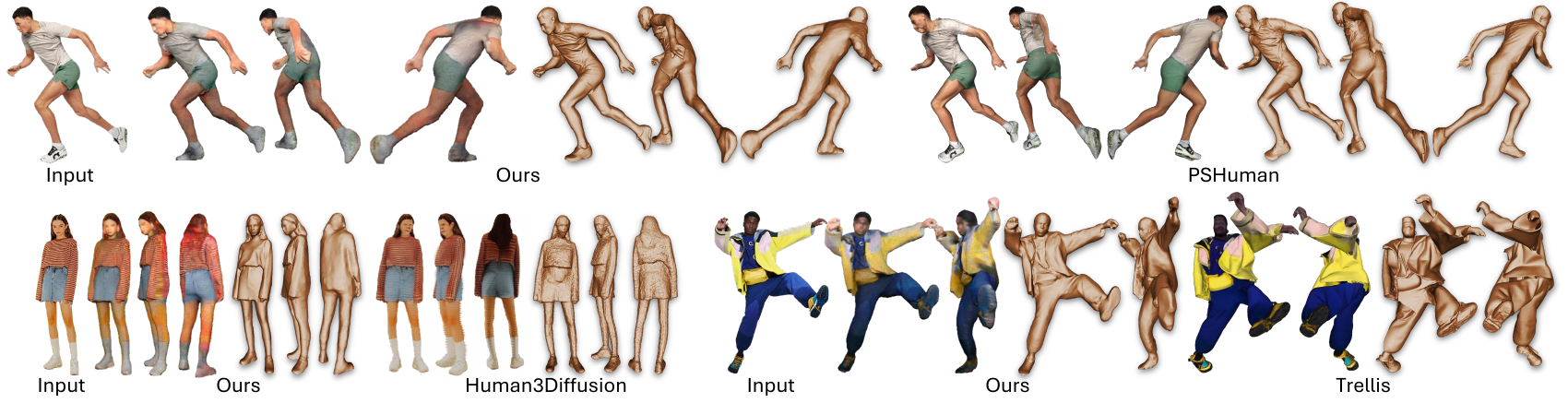}
	\caption{Geometry and appearance comparisons of our method against PSHuman \cite{li2025pshuman}, Human3Diffusion \cite{xue2024human} and Trellis \cite{xiang2025structured} on in-the-wild images. \textcolor{red}{\faSearch} Zoom in for details.}
	\label{wildcompare}
\end{figure*}

\subsection{Implementation Details} 
The sparse VAE was trained using Minkowski Engine 0.5.4 with a batch size of 8 for 200,000 iterations. The Adam optimizer was employed with a learning rate of 0.00035. The loss weights $\lambda_1 \sim \lambda_7$ were set as 0.8, 0.2, 0.1, $5\times 10^{-7}$, 1, 1, 1, respectively. (To improve training stability, $\mathcal{L}_{\text{Attr}}$ and $\mathcal{L}_{\text{Render}}$ were introduced only after 10,000 iterations.) For bridge diffusion, both training and inference followed the original DDBM setting \cite{zhou2023denoising}, where the parameters \texttt{CHURN\_STEP\_RATIO} and \texttt{GUIDANCE} were set to 0.1 and 1, respectively. The batch size was set to 16, and the number of training iterations was 100,000. The Adam optimizer was employed with a learning rate of 0.00035. Depth Anything V2 \cite{yang2024depth} was adopted as the backbone for depth estimation and PIXIE \cite{feng2021collaborative} was selected to predict the SMPL models. The sparse 3D UNet was trained with a batch size of 8 for 40,000 iterations. The Adam optimizer was employed with a learning rate of 0.00035. All training and testing were conducted on a server equipped with four NVIDIA A6000 GPUs.

The parameter counts of the sparse VAE, bridge-diffusion 3D U-Net, and sparse U-Net are 9.04M, 229.76M, and 36.99M, respectively. The training times for these models are 2 days, 4 days, and 1 day, respectively. The GPU memory cost for these models are 40 GB, 4$\times$45 GB, and 30 GB, respectively. The inference time for a single sample is approximately 2 minutes.

\subsection{Comparison with State-of-the-art Methods}
We mainly compare our JGA-LBD with three representative 3DGS-based approaches. IDOL \cite{zhang2025idol} leverages an explicit SMPL model as a geometry prior to guide 3D Gaussians generation. MultiGo \cite{zhang2025multigo} generates a complete 3D Gaussian scene using a large model, while encoding SMPL as Fourier features to provide structural guidance. Trellis \cite{xiang2025structured} learns a compact latent space that jointly encodes both geometry and appearance for structured generative modeling with multiple stages. We further compare JGA-LBD with several recent mesh-based methods \cite{ho2024sith,zhang2023global,zhang2024sifu,li2025pshuman}. As shown in Table~\ref{sotaresults}, JGA-LBD achieves the best performance across all metrics on both benchmark datasets.

\vspace{0.5em}
\noindent\textbf{Geometry Comparison.} Human3Diffusion \cite{xue2024human} produces noticeably noisy geometry and fails to preserve fine frontal wrinkle details. IDOL \cite{zhang2025idol} heavily relies on the SMPL model without any refinement, and therefore, as shown in Figure~\ref{Geometrycompare}, it often produces incorrect poses. Moreover, due to the strong regularization imposed by SMPL, it fails to handle loose clothing such as dresses (see the third case).  Although MultiGo \cite{zhang2025multigo} employs a wrinkle refinement network, its geometric reconstruction still lacks fine details. Moreover, as observed in the third and fourth cases in Figure~\ref{Geometrycompare}, the reconstructed bodies exhibit a forward-leaning tendency. This indicates that, although MultiGo avoids the pose inaccuracies introduced by directly using SMPL, its 2D diffusion model is insufficient to correct pose errors in 3D space. Trellis \cite{xiang2025structured} suffers from low mesh resolution, which severely limits the reconstruction of fine details. In addition, the reconstructed poses are often inaccurate, with head rotations consistently misaligned with the input across all cases. In contrast, our JGA-LBD is able to reconstruct fine geometric details and handle loose clothing, while maintaining accurate overall human poses. \\

\vspace{0.5em}
\noindent\textbf{Appearance Comparison.} Human3Diffusion \cite{xue2024human} shows many jagged artifacts in the rendered images, and it cannot recover texture details accurately. IDOL \cite{zhang2025idol} suffers from severe misalignment caused by wrong SMPL poses. As shown in the Figure. \ref{visualcompare}, it can only capture relatively simple color patterns and fails to represent fine-grained textures such as stripes in the second case. In addition, noticeable jagged artifacts can be observed along the edges. MultiGo \cite{zhang2025multigo} performs well on the front side, but its back-side reconstructions remain poor. For example, in the second case it fails to recover the stripe patterns, and in the third case the back of the head incorrectly contains facial details instead of black hair. Beyond its failure to reconstruct fine details such as stripes, Trellis \cite{xiang2025structured} also suffers from severe geometry–appearance inconsistencies. For instance, in the second and third cases, the reconstructed arms are noticeably inconsistent with those shown in Fig. \ref{Geometrycompare}. In contrast, our JGA-LBD not only reconstructs fine details such as stripes and back-side wrinkles, but also maintains geometric consistency with the reference in Fig. \ref{Geometrycompare}.

\vspace{0.5em}
\noindent\textbf{Comparison with PSHuman \cite{li2025pshuman}.} We also compare our JGA-LBD with the popular work of PSHuman. As shown in Fig. \ref{pshumancompare}, PSHuman frequently suffers from missing body parts. PSHuman also struggles with handling the relative positions of the legs in the first example. In the second example, it also fails to preserve facial details, producing noticeably degraded results. In contrast, our method effectively avoids these issues.  \\

Overall, both quantitative metrics and qualitative comparisons strongly demonstrate that our JGA-LBD consistently outperforms current state-of-the-art methods. 
We further conducted experiments on in-the-wild images with challenging poses, as shown in Fig.~\ref{inthewild}, where JGA-LBD successfully reconstructs plausible appearances and detailed 3D surfaces. We also compare our method with representative approaches, including the mesh-based PSHuman \cite{li2024pshuman}, the 3DGS-based Human3Diffusion \cite{xue2024human}, and the general-purpose method Trellis \cite{xiang2025structured}, on in-the-wild images, as shown in Fig.~\ref{wildcompare}.
PSHuman \cite{li2024pshuman} fails to preserve fine facial details and produces inaccurate hand reconstructions. Human3Diffusion \cite{xue2024human} struggles to generate correct human poses and exhibits degraded surface quality and appearance details. Trellis \cite{xiang2025structured}, as a general reconstruction framework, often fails to produce a structurally plausible human shape. In contrast, our method achieves more accurate geometry and more faithful appearance reconstruction under complex real-world conditions. We refer the reader to the \textbf{video demo} attached.

\begin{table}[t]
    \centering
    \caption{Ablation studies on 2K2K. }
    \label{ablation}
    \scriptsize
    \resizebox{\linewidth}{!}{\begin{tabular}{l|c|c|c|c|c|c} 
			\toprule
			\multicolumn{1}{c|}{{\diagbox{Method}{Metric}}}    
			& \multicolumn{1}{c}{PSNR$\uparrow$} & \multicolumn{1}{c}{SSIM$\uparrow$} & \multicolumn{1}{c}{LPIPS$\downarrow$} & \multicolumn{1}{c}{CD$\downarrow$} & \multicolumn{1}{c}{P2S$\downarrow$} & \multicolumn{1}{c}{Normal$\downarrow$}  \\
			\hline
			\hline
			Rectified Flow  & 28.32  & 0.931  & 0.074 & 0.568 & 0.535 & 1.436  \\ 
            Feature Condition  & Fail
			& Fail & Fail & Fail & Fail & Fail  \\
            \hline
              w/o Decode Module & 27.68 &0.931&0.073 & 0.498 & \textbf{0.497} & 1.287  \\  
			\hline
            EcoDepth + PIXIE &30.01 & 0.944 & 0.059 & 0.536&0.544&1.288\\
            DepthAnything + PyMAF &29.51 & 0.943 & 0.059  & 0.540&0.543&1.367\\
            \hline
			Full Model & \textbf{30.16} & \textbf{0.946} & \textbf{0.055} & \textbf{0.489} & 0.507 & \textbf{1.202}  \\ 
			\bottomrule
 	\end{tabular}}
\end{table}

\begin{figure}[t]
	\centering
	\includegraphics[width=3.0in]{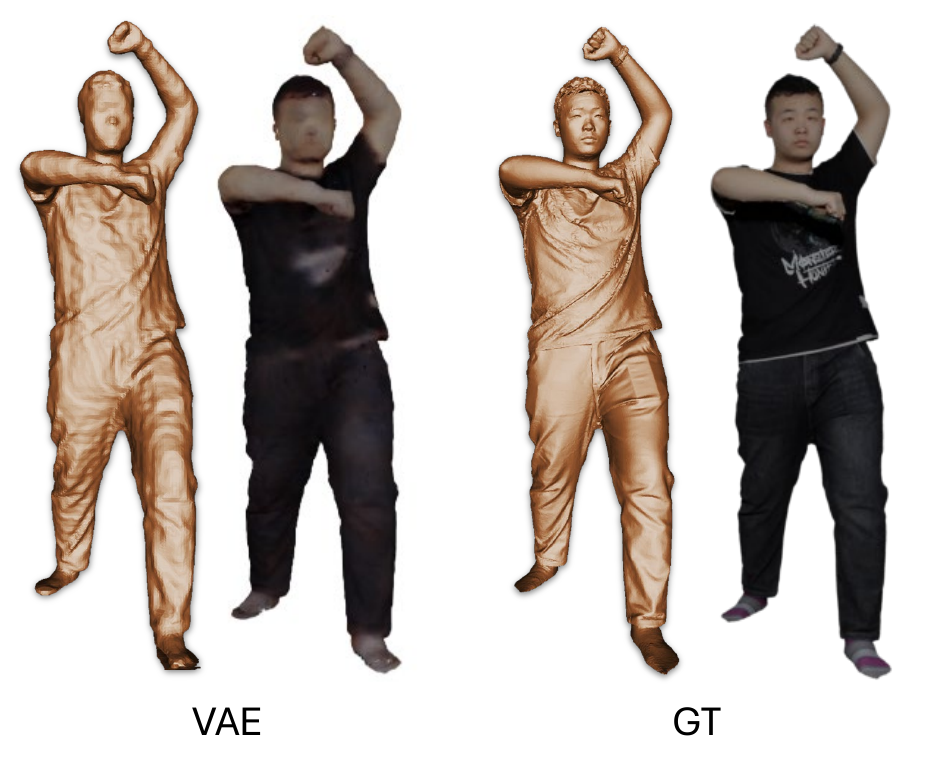}
	\caption{ The visualization comparison between VAE results and GT. \textcolor{red}{\faSearch} Zoom in for details.}
	\label{ablationfigurea}
\end{figure}

\subsection{Ablation Studies}

\noindent \textbf{Visual Results of Sparse VAE.} To evaluate the effectiveness of the sparse VAE, we conduct experiments on reconstructing human 3D Gaussian representations directly from the compressed latent space. Specifically, the input 3D Gaussians are compressed from $(512^3, 20)$ into a latent tensor of size $(64^3, 4)$, and subsequently decoded to recover 3D Gaussian attributes. As shown in Fig.~\ref{ablationfigurea}, the sparse VAE is able to preserve the overall geometry and coarse appearance of the human body, demonstrating that the latent space effectively encodes both structural and visual information. However, fine-grained details such as sharp geometric boundaries and high-frequency textures are noticeably degraded due to the high compression ratio. This observation motivates the introduction of a decode module after the VAE decoder to enhance reconstruction fidelity. 

\begin{figure}[t]
	\centering
	\includegraphics[width=3.0in]{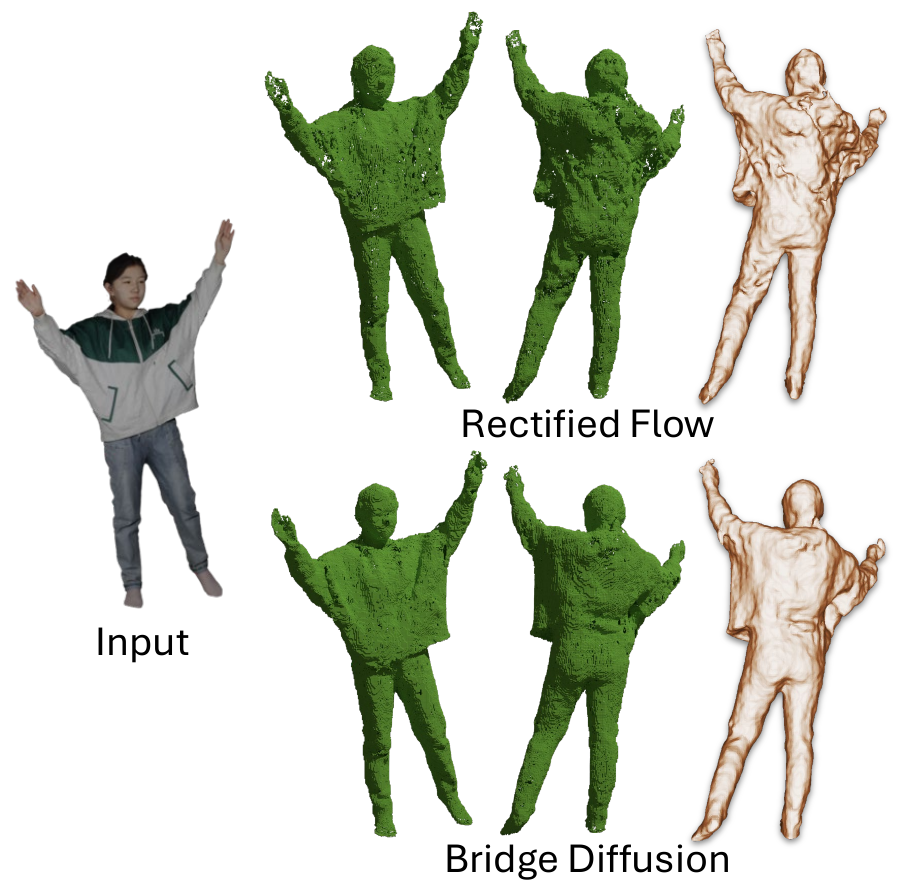}
	\caption{ Comparison between rectified flow and bridge diffusion. \textcolor{red}{\faSearch} Zoom in for details.}
	\label{ablationfigureb}
\end{figure}

\vspace{0.5em}
\noindent\textbf{Effectiveness of Bridge Diffusion.} We further compare the bridge diffusion employed in our work with the popular rectified flow method \cite{liu2022flow}, the quantitative results are shown in Table \ref{ablation}. The visual results in Fig. \ref{ablationfigureb} show that rectified flow tends to generate 3D Gaussians with many holes, and the reconstructed back surfaces are heavily corrupted by noise. This observation highlights the advantage of bridge diffusion: since the starting depth is already part of the complete 3D Gaussians, bridge diffusion does not need to allocate excessive capacity to the visible front side but instead focuses on learning the missing regions. As a result, our strategy of adopting bridge diffusion achieves superior reconstruction quality.

\begin{figure}[t]
	\centering
	\includegraphics[width=3.0in]{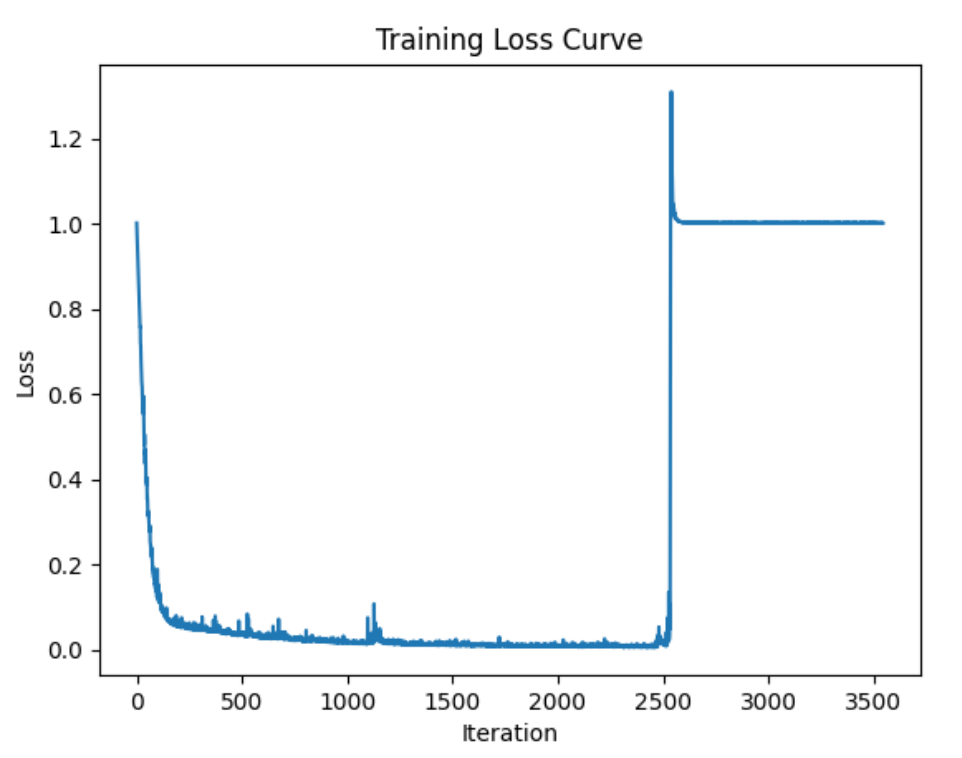}
	\caption{ Training loss curve of image and SMPL feature level supervision. \textcolor{red}{\faSearch} Zoom in for details.}
	\label{ablationfigurec}
\end{figure}

\vspace{0.5em}
\noindent\textbf{Necessity of Structural Prior.} We further validate the necessity of introducing structural priors. Following GaussianCube \cite{zhanggaussiancube}, we extract image features using DINOv2 \cite{oquab2023dinov2} and SMPL features using Point-M2AE \cite{zhang2022point}, and employ both features to supervise the training process of the bridge diffusion model. However, the training often fails to converge: as shown in Fig. \ref{ablationfigurec}, the loss decreases at the beginning but suddenly rises after several epochs, eventually leading to divergence. We attribute this to the modality gap between images and SMPL, which makes it difficult for the model to learn meaningful representations when directly using such heterogeneous features. This observation underscores the importance of our unified latent space, where features from different modalities are mapped into a shared representation, substantially reducing training difficulty and improving stability. 

\begin{figure}[t]
	\centering
	\includegraphics[width=3.3in]{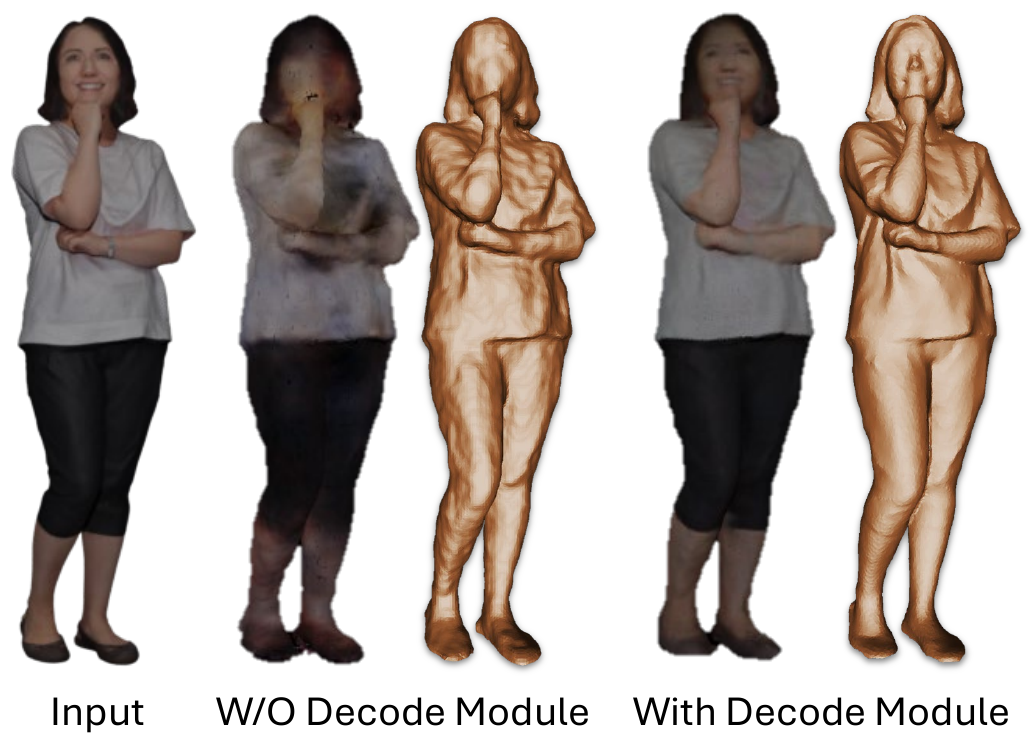}
	\caption{ Comparison between visual results without and with the decode module. \textcolor{red}{\faSearch} Zoom in for details.}
	\label{ablationfigured}
\end{figure}

\vspace{0.5em}
\noindent \textbf{Necessity of Decode Module.} Due to the aggressive compression ratio of the VAE, inevitable information loss is introduced. In addition, the diffusion model itself cannot achieve perfectly error-free reconstruction, and such residual errors further amplify the loss caused by compression, making it difficult to capture and recover high-frequency details. To address this issue, we introduce a decoding module after the diffusion model to refine and enhance the generated results. As shown in Table \ref{ablation}, the decode module brings significant improvement on both appearance and geometry metrics. Fig. \ref{ablationfigured}, we can also observe enhanced details in both appearance and geometry. Moreover, it is worth noting that even without the additional decode module, the results already achieve the best performance in terms of geometry and deliver appearance reconstruction that remains competitive with state-of-the-art methods. 
To quantify the effect of different depth estimation methods, we compare two depth estimators—Depth Anything V2 \cite{yang2024depth} (RMSE 0.014 on 2K2K) and EcoDepth \cite{patni2024ecodepth} (RMSE 0.016 on 2K2K)—and observe that better depth leads to consistently better reconstruction performance (Table~\ref{ablation}). We also evaluate different SMPL regressors (i.e., PIXIE \cite{feng2021collaborative} vs. PyMAF \cite{zhang2021pymaf}) and find that they produce comparable results with minor variations.

\begin{figure}[t]
	\centering
	\includegraphics[width=3.3in]{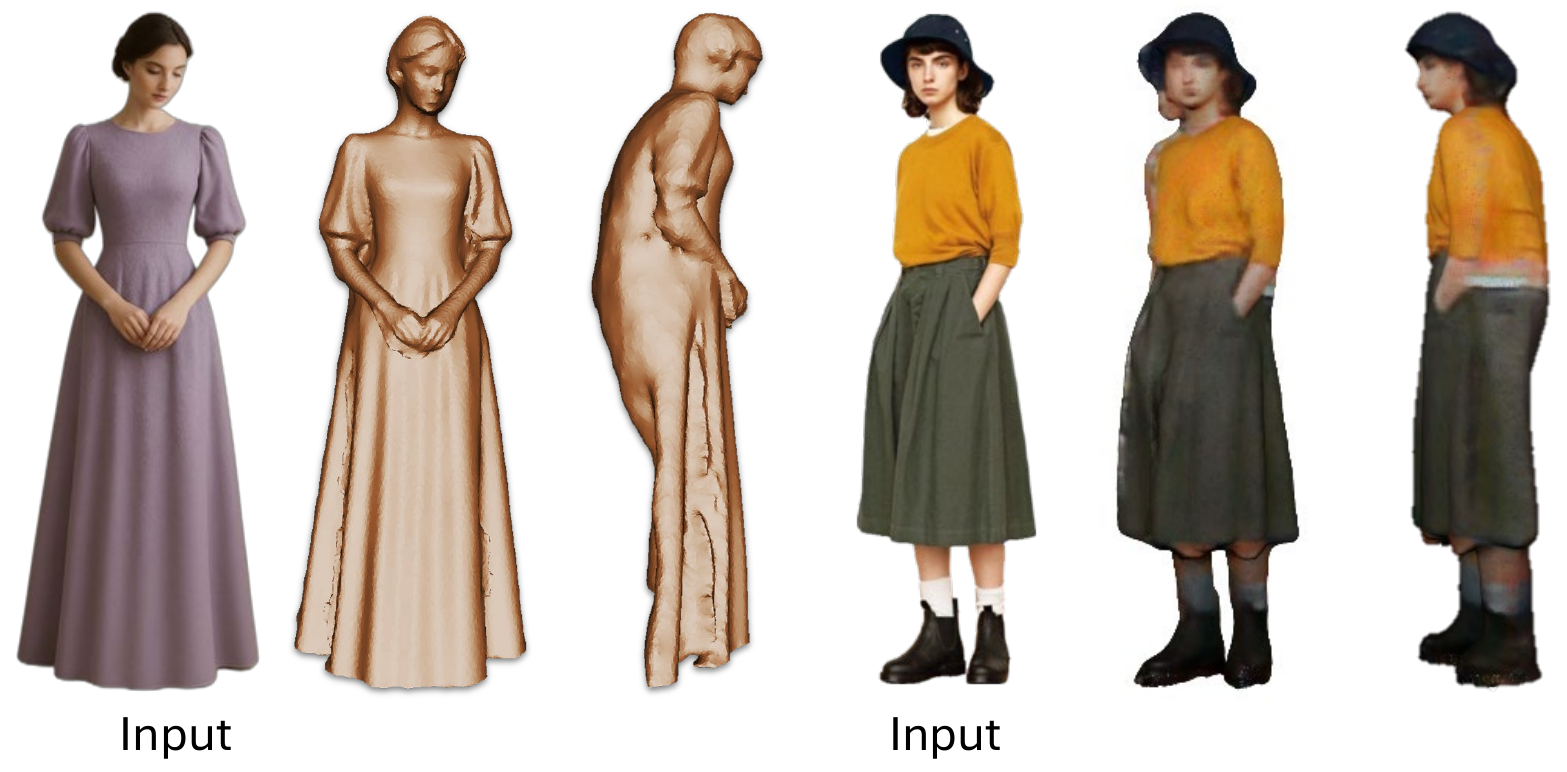}
	\caption{Illustration of the limitations of JGA-LBD.}
	\label{failurecase}
\end{figure}

\vspace{0.5em}
 \subsection{Limitations}
We provide two representative types of failure cases in Fig. \ref{failurecase}.
The first limitation arises from the limited availability of high-quality 3D human scan data. Due to the scarcity of diverse training samples, the learned latent space lacks sufficient coverage of complex clothing distributions, which restricts the model’s generalization ability. As illustrated in Fig. \ref{failurecase}, although our method successfully reconstructs a plausible human body and produces visually reasonable results from the front view, it fails to recover the correct volumetric structure of loose garments such as long dresses. Specifically, the back of the dress should exhibit a voluminous shape, whereas the reconstructed result incorrectly collapses toward the legs. This limitation stems from the insufficient representation of such clothing styles in the training data, preventing the model from learning their characteristic geometric distributions.
The second type of failure concerns appearance reconstruction. In some cases, the model fails to faithfully recover the frontal color information that is explicitly visible in the input image. We attribute this issue primarily to the limited scale and diversity of the training data. Specifically, the scarcity of high-quality 3D human scans constrains the representational capacity of the learned VAE latent space, which in turn limits its ability to generalize to complex scenes or uncommon appearance configurations, ultimately affecting the performance of the subsequent diffusion model.

\section{Conclusion}
\label{sectionconclusion}
In this work, we have presented JGA-LBD, a framework that reconstructs both geometry and appearance of a human in a single generation step. Experimental results demonstrate that our method achieves superior performance in both geometry and appearance reconstruction compared to state-of-the-art methods, and performs well on in-the-wild images with hard poses and loose clothing. Unlike existing methods that decouple geometry and appearance, JGA-LBD performs joint modeling, thereby ensuring better consistency between geometry and appearance.

In future work, we will exploit a more powerful sparse VAE capable of capturing high-frequency details and explore diffusion architectures that eliminate the need for an additional refinement decoder, further improving both efficiency and reconstruction quality. Besides, we will investigate more effective strategies for learning a discriminative and robust latent space from limited 3D human scans, with the goal of better supporting diffusion-based generation under challenging real-world conditions.

{
\bibliographystyle{ieee_fullname}
\bibliography{egbib}

@inproceedings{ho2023learning,
  title     = {Learning locally editable virtual humans},
  author    = {Ho, Hsuan-I and Xue, Lixin and Song, Jie and Hilliges, Otmar},
  booktitle = { IEEE/CVF Conference on Computer Vision and Pattern Recognition},
    pages={21024--21035},
  year      = 2023
}

@article{zhang2022point,
  title={Point-m2ae: multi-scale masked autoencoders for hierarchical point cloud pre-training},
  author={Zhang, Renrui and Guo, Ziyu and Gao, Peng and Fang, Rongyao and Zhao, Bin and Wang, Dong and Qiao, Yu and Li, Hongsheng},
  journal={Advances in Neural Information Processing Systems},
  volume={35},
  pages={27061--27074},
  year={2022}
}

@article{palfinger2022continuous,
  title={Continuous remeshing for inverse rendering},
  author={Palfinger, Werner},
  journal={Computer Animation and Virtual Worlds},
  volume={33},
  number={5},
  pages={e2101},
  year={2022}
}

@inproceedings{xue2024human,
  title     = {Human 3Diffusion: Realistic Avatar Creation via Explicit 3D Consistent Diffusion Models},
  author    = {Xue, Yuxuan and Xie, Xianghui and Marin, Riccardo and Pons-Moll, Gerard},
  booktitle = { Advances in Neural Information Processing Systems},
  year      = 2024
}

@inproceedings{zhanggaussiancube,
  title     = {GaussianCube: A Structured and Explicit Radiance Representation for 3D Generative Modeling},
  author    = {Zhang, Bowen and Cheng, Yiji and Yang, Jiaolong and Wang, Chunyu and Zhao, Feng and Tang, Yansong and Chen, Dong and Guo, Baining},
  booktitle = { Advances in Neural Information Processing Systems},
  year      = 2024
}

@inproceedings{ho2024sith,
  title     = {Sith: Single-view textured human reconstruction with image-conditioned diffusion},
  author    = {Ho, I and Song, Jie and Hilliges, Otmar and others},
  booktitle = { IEEE/CVF Conference on Computer Vision and Pattern Recognition},
    pages={538--549},
  year      = 2024
}

@inproceedings{choy20194d,
  title={4D Spatio-Temporal ConvNets: Minkowski Convolutional Neural Networks},
  author={Choy, Christopher and Gwak, JunYoung and Savarese, Silvio},
  booktitle={IEEE/CVF Conference on Computer Vision and Pattern Recognition},
  pages={3075--3084},
  year={2019}
}

@inproceedings{liu2023zero,
  title     = {Zero-1-to-3: Zero-shot one image to 3d object},
  author    = {Liu, Ruoshi and Wu, Rundi and Van Hoorick, Basile and Tokmakov, Pavel and Zakharov, Sergey and Vondrick, Carl},
  booktitle = { IEEE/CVF International Conference on Computer Vision},
    pages={9298--9309},
  year      = 2023
}

@inproceedings{feng2021collaborative,
  title={Collaborative regression of expressive bodies using moderation},
  author={Feng, Yao and Choutas, Vasileios and Bolkart, Timo and Tzionas, Dimitrios and Black, Michael J},
  booktitle={International Conference on 3D Vision},
  pages={792--804},
  year={2021}
}

@article{yang2024depth,
  title={Depth anything v2},
  author={Yang, Lihe and Kang, Bingyi and Huang, Zilong and Zhao, Zhen and Xu, Xiaogang and Feng, Jiashi and Zhao, Hengshuang},
  journal={Advances in Neural Information Processing Systems},
  volume={37},
  pages={21875--21911},
  year={2024}
}

@article{oquab2023dinov2,
  title={Dinov2: Learning robust visual features without supervision},
  author={Oquab, Maxime and Darcet, Timoth{\'e}e and Moutakanni, Th{\'e}o and Vo, Huy and Szafraniec, Marc and Khalidov, Vasil and Fernandez, Pierre and Haziza, Daniel and Massa, Francisco and El-Nouby, Alaaeldin and others},
  journal={arXiv preprint arXiv:2304.07193},
  year={2023}
}

@article{tang2025human,
  title={Human as points: Explicit point-based 3d human reconstruction from single-view rgb images},
  author={Tang, Yingzhi and Zhang, Qijian and Liu, Yebin and Hou, Junhui},
  journal={ IEEE Transactions on Pattern Analysis and Machine Intelligence},
  year={2025}
}

@inproceedings{luo2021diffusion,
  title     = {Diffusion probabilistic models for 3d point cloud generation},
  author    = {Luo, Shitong and Hu, Wei},
  booktitle = { IEEE/CVF Conference on Computer Vision and Pattern Recognition},
    pages={2837--2845},
  year      = 2021
}

@inproceedings{mildenhall2020nerf,
  title     = {{NeRF}: Representing Scenes as Neural Radiance Fields for View Synthesis},
  author    = {Mildenhall, Ben and Srinivasan, Pratul P and Tancik, Matthew and Barron, Jonathan T and Ramamoorthi, Ravi and Ng, Ren},
  booktitle = { European Conference on Computer Vision},
    pages={99--106},
  year      = 2020
}

@article{kerbl20233d,
  title     = {3D Gaussian Splatting for Real-Time Radiance Field Rendering.},
  author    = {Kerbl, Bernhard and Kopanas, Georgios and Leimk{\"u}hler, Thomas and Drettakis, George},
  journal   = {ACM Transactions on Graphics},
  volume    = 42,
  number    = 4,
  pages     = {139--1},
  year      = 2023
}

@article{ho2022classifier,
  title={Classifier-free diffusion guidance},
  author={Ho, Jonathan and Salimans, Tim},
  journal={Advances in Neural Information Processing Systems},
  year={2021}
}

@article{li2024craftsman3d,
  title={Craftsman3d: High-fidelity mesh generation with 3d native generation and interactive geometry refiner},
  author={Li, Weiyu and Liu, Jiarui and Yan, Hongyu and Chen, Rui and Liang, Yixun and Chen, Xuelin and Tan, Ping and Long, Xiaoxiao},
  journal={IEEE/CVF Conference on Computer Vision and Pattern Recognition},
  year={2025}
}

@article{qiu2025lhm,
  title={Lhm: Large animatable human reconstruction model from a single image in seconds},
  author={Qiu, Lingteng and Gu, Xiaodong and Li, Peihao and Zuo, Qi and Shen, Weichao and Zhang, Junfei and Qiu, Kejie and Yuan, Weihao and Chen, Guanying and Dong, Zilong and others},
  journal={IEEE/CVF International Conference on Computer Vision},
  year={2025}
}

@InProceedings{2k2k,
    author    = {Han, Sang-Hun and Park, Min-Gyu and Yoon, Ju Hong and Kang, Ju-Mi and Park, Young-Jae and Jeon, Hae-Gon},
    title     = {{High-fidelity 3d human digitization from single 2k resolution images}},
    booktitle = {IEEE/CVF Computer Vision and Pattern Recognition Conference},
    pages     = {12869-12879},
    year={2023}
}

@inproceedings{li2024pshuman,
  title={PSHuman: Photorealistic Single-view Human Reconstruction using Cross-Scale Diffusion},
  author={Li, Peng and Zheng, Wangguandong and Liu, Yuan and Yu, Tao and Li, Yangguang and Qi, Xingqun and Li, Mengfei and Chi, Xiaowei and Xia, Siyu and Xue, Wei and others},
  booktitle={IEEE/CVF Computer Vision and Pattern Recognition Conference},
  year={2024}
}

@article{tang2023high,
  title={High-resolution volumetric reconstruction for clothed humans},
  author={Tang, Sicong and Wang, Guangyuan and Ran, Qing and Li, Lingzhi and Shen, Li and Tan, Ping},
  journal={ACM Transactions on Graphics},
  year={2023}
}

@article{xiu2022econ,
  title={{ECON: Explicit Clothed humans Obtained from Normals}},
  author={Xiu, Yuliang and Yang, Jinlong and Cao, Xu and Tzionas, Dimitrios and Black, Michael J},
  journal={  IEEE/CVF Computer Vision and Pattern Recognition Conference},
  year={2023}
}

@inproceedings{zhang2021pymaf,
  title={{PyMAF}: 3d human pose and shape regression with pyramidal mesh alignment feedback loop},
  author={Zhang, Hongwen and Tian, Yating and Zhou, Xinchi and Ouyang, Wanli and Liu, Yebin and Wang, Limin and Sun, Zhenan},
  booktitle={ IEEE/CVF International Conference on Computer Vision},
  pages={11446--11456},
  year={2021}
}

@inproceedings{patni2024ecodepth,
  title={{ECoDepth}: Effective Conditioning of Diffusion Models for Monocular Depth Estimation},
  author={Patni, Suraj and Agarwal, Aradhye and Arora, Chetan},
  booktitle={ IEEE/CVF Computer Vision and Pattern Recognition Conference },
  pages={28285--28295},
  year={2024}
}

@inproceedings{ren2024xcube,
  title={Xcube: Large-scale 3d generative modeling using sparse voxel hierarchies},
  author={Ren, Xuanchi and Huang, Jiahui and Zeng, Xiaohui and Museth, Ken and Fidler, Sanja and Williams, Francis},
  booktitle={IEEE/CVF Computer Vision and Pattern Recognition Conference},
  pages={4209--4219},
  year={2024}
}

@inproceedings{natsume2019siclope,
  title={{SiCloPe}: Silhouette-Based Clothed People},
  author={Natsume, Ryota and Saito, Shunsuke and Huang, Zeng and Chen, Weikai and Ma, Chongyang and Li, Hao and Morishima, Shigeo},
  booktitle={ IEEE/CVF Computer Vision and Pattern Recognition Conference },
  pages={4480--4490},
  year={2019}
}

@article{kazhdan2013screened,
  title={Screened poisson surface reconstruction},
  author={Kazhdan, Michael and Hoppe, Hugues},
  journal={ACM Transactions on Graphics},
  volume={32},
  pages={1--13},
  year={2013}
}

@inproceedings{yu2021function4d,
  title     = {Function4d: Real-time human volumetric capture from very sparse consumer rgbd sensors},
  author    = {Yu, Tao and Zheng, Zerong and Guo, Kaiwen and Liu, Pengpeng and Dai, Qionghai and Liu, Yebin},
  booktitle = { IEEE/CVF Conference on Computer Vision and Pattern Recognition},
  pages = {5746--5756},
  year      = 2021
}

@inproceedings{li2025pshuman,
  title={Pshuman: Photorealistic single-image 3d human reconstruction using cross-scale multiview diffusion and explicit remeshing},
  author={Li, Peng and Zheng, Wangguandong and Liu, Yuan and Yu, Tao and Li, Yangguang and Qi, Xingqun and Chi, Xiaowei and Xia, Siyu and Cao, Yan-Pei and Xue, Wei and others},
  booktitle={ IEEE/CVF Conference on Computer Vision and Pattern Recognition},
  pages={16008--16018},
  year={2025}
}

@inproceedings{tang2024lgm,
  title     = {Lgm: Large multi-view gaussian model for high-resolution 3d content creation},
  author    = {Tang, Jiaxiang and Chen, Zhaoxi and Chen, Xiaokang and Wang, Tengfei and Zeng, Gang and Liu, Ziwei},
  booktitle = { European Conference on Computer Vision},
  pages={1--18},
  year      = 2024
}

@inproceedings{zhou2024diffgs,
  title     = {DiffGS: Functional gaussian splatting diffusion},
  author    = {Zhou, Junsheng and Zhang, Weiqi and Liu, Yu-Shen},
  booktitle = { Advances in Neural Information Processing Systems},
  year      = 2024
}

@inproceedings{zhang2023global,
  title     = {Global-correlated 3d-decoupling transformer for clothed avatar reconstruction},
  author    = {Zhang, Zechuan and Sun, Li and Yang, Zongxin and Chen, Ling and Yang, Yi},
  booktitle = { Advances in Neural Information Processing Systems},
  year      = 2023
}

@inproceedings{zhang2024sifu,
  title     = {Sifu: Side-view conditioned implicit function for real-world usable clothed human reconstruction},
  author    = {Zhang, Zechuan and Yang, Zongxin and Yang, Yi},
  booktitle = { IEEE/CVF Conference on Computer Vision and Pattern Recognition},
  pages={9936--9947},
  year      = 2024
}

@inproceedings{huang2024tech,
  title={Tech: Text-guided reconstruction of lifelike clothed humans},
  author={Huang, Yangyi and Yi, Hongwei and Xiu, Yuliang and Liao, Tingting and Tang, Jiaxiang and Cai, Deng and Thies, Justus},
  booktitle={International Conference on 3D Vision},
  pages={1531--1542},
  year={2024}
}

@article{chen2024generalizable,
  title={Generalizable human gaussians from single-view image},
  author={Chen, Jinnan and Li, Chen and Zhang, Jianfeng and Zhu, Lingting and Huang, Buzhen and Chen, Hanlin and Lee, Gim Hee},
  journal={International Conference on Learning Representations},
  year={2025},
}

@article{pan2024humansplat,
  title={Humansplat: Generalizable single-image human gaussian splatting with structure priors},
  author={Pan, Panwang and Su, Zhuo and Lin, Chenguo and Fan, Zhen and Zhang, Yongjie and Li, Zeming and Shen, Tingting and Mu, Yadong and Liu, Yebin},
  journal={Advances in Neural Information Processing Systems},
  volume={37},
  pages={74383--74410},
  year={2024}
}

@article{wang2025wonderhuman,
  title={Wonderhuman: Hallucinating unseen parts in dynamic 3d human reconstruction},
  author={Wang, Zilong and Dou, Zhiyang and Liu, Yuan and Lin, Cheng and Dong, Xiao and Guo, Yunhui and Zhang, Chenxu and Li, Xin and Wang, Wenping and Guo, Xiaohu},
  journal={IEEE Transactions on Visualization and Computer Graphics},
  year={2025}
}

@article{hu2025humangif,
  title={Humangif: Single-view human diffusion with generative prior},
  author={Hu, Shoukang and Narihira, Takuya and Fukuda, Kazumi and Sawata, Ryosuke and Shibuya, Takashi and Mitsufuji, Yuki},
  journal={arXiv preprint arXiv:2502.12080},
  year={2025}
}

@article{wang2024geneman,
  title={Geneman: Generalizable single-image 3d human reconstruction from multi-source human data},
  author={Wang, Wentao and Ye, Hang and Hong, Fangzhou and Yang, Xue and Zhang, Jianfu and Wang, Yizhou and Liu, Ziwei and Pan, Liang},
   booktitle = { Advances in Neural Information Processing Systems},
  year      = 2025
}

@inproceedings{he2025magicman,
  title={Magicman: Generative novel view synthesis of humans with 3d-aware diffusion and iterative refinement},
  author={He, Xu and Wu, Zhiyong and Li, Xiaoyu and Kang, Di and Zhang, Chaopeng and Ye, Jiangnan and Chen, Liyang and Gao, Xiangjun and Zhang, Han and Zhuang, Haolin},
  booktitle={AAAI},
  volume={39},
  number={3},
  pages={3437--3445},
  year={2025}
}

@inproceedings{albahar2023single,
  title={Single-image 3d human digitization with shape-guided diffusion},
  author={AlBahar, Badour and Saito, Shunsuke and Tseng, Hung-Yu and Kim, Changil and Kopf, Johannes and Huang, Jia-Bin},
  booktitle={SIGGRAPH Asia 2023 Conference Papers},
  pages={1--11},
  year={2023}
}

@article{tang2025hugdiffusion,
  title={HuGDiffusion: Generalizable Single-Image Human Rendering via 3D Gaussian Diffusion},
  author={Tang, Yingzhi and Zhang, Qijian and Hou, Junhui},
  journal   = { IEEE Transactions on Visualization and Computer Graphics},
  year      = 2025
}

@inproceedings{rombach2022high,
  title     = {High-resolution image synthesis with latent diffusion models},
  author    = {Rombach, Robin and Blattmann, Andreas and Lorenz, Dominik and Esser, Patrick and Ommer, Bj{\"o}rn},
  booktitle = { IEEE/CVF Conference on Computer Vision and Pattern Recognition},
  pages={10684--10695},
  year      = 2022
}

@inproceedings{zhang2023adding,
  title     = {Adding conditional control to text-to-image diffusion models},
  author    = {Zhang, Lvmin and Rao, Anyi and Agrawala, Maneesh},
  booktitle = { IEEE/CVF International Conference on Computer Vision},
  pages={3836--3847},
  year      = 2023
}

@inproceedings{saito2019pifu,
  title     = {Pifu: Pixel-aligned implicit function for high-resolution clothed human digitization},
  author    = {Saito, Shunsuke and Huang, Zeng and Natsume, Ryota and Morishima, Shigeo and Kanazawa, Angjoo and Li, Hao},
  booktitle = { IEEE/CVF International Conference on Computer Vision},
    pages={2304--2314},
  year      = 2019
}

@article{ho2020denoising,
  title={Denoising diffusion probabilistic models},
  author={Ho, Jonathan and Jain, Ajay and Abbeel, Pieter},
  journal={Advances in Neural Information Processing Systems},
  volume={33},
  pages={6840--6851},
  year={2020}
}

@inproceedings{xiang2025structured,
  title={Structured 3d latents for scalable and versatile 3d generation},
  author={Xiang, Jianfeng and Lv, Zelong and Xu, Sicheng and Deng, Yu and Wang, Ruicheng and Zhang, Bowen and Chen, Dong and Tong, Xin and Yang, Jiaolong},
  booktitle={IEEE/CVF Conference on Computer Vision and Pattern Recognition},
  pages={21469--21480},
  year={2025}
}

@article{zhang20233dshape2vecset,
  title={3dshape2vecset: A 3d shape representation for neural fields and generative diffusion models},
  author={Zhang, Biao and Tang, Jiapeng and Niessner, Matthias and Wonka, Peter},
  journal={ACM Transactions on Graphics},
  volume={42},
  number={4},
  pages={1--16},
  year={2023}
}

@inproceedings{wang2023score,
  title={Score jacobian chaining: Lifting pretrained 2d diffusion models for 3d generation},
  author={Wang, Haochen and Du, Xiaodan and Li, Jiahao and Yeh, Raymond A and Shakhnarovich, Greg},
  booktitle={IEEE/CVF Conference on Computer Vision and Pattern Recognition},
  pages={12619--12629},
  year={2023}
}

@article{tang2023dreamgaussian,
  title={Dreamgaussian: Generative gaussian splatting for efficient 3d content creation},
  author={Tang, Jiaxiang and Ren, Jiawei and Zhou, Hang and Liu, Ziwei and Zeng, Gang},
  journal={International Conference on Learning Representations},
  year={2024}
}

@article{dhariwal2021diffusion,
  title={Diffusion models beat gans on image synthesis},
  author={Dhariwal, Prafulla and Nichol, Alexander},
  journal={Advances in Neural Information Processing Systems},
  volume={34},
  pages={8780--8794},
  year={2021}
}

@inproceedings{li2023bbdm,
  title={Bbdm: Image-to-image translation with brownian bridge diffusion models},
  author={Li, Bo and Xue, Kaitao and Liu, Bin and Lai, Yu-Kun},
  booktitle={IEEE/CVF Conference on Computer Vision and Pattern Recognition},
  pages={1952--1961},
  year={2023}
}

@article{zhou2023denoising,
  title={Denoising diffusion bridge models},
  author={Zhou, Linqi and Lou, Aaron and Khanna, Samar and Ermon, Stefano},
  journal={International Conference on Learning Representations},
  year={2024}
}

@article{liu2022flow,
  title={Flow straight and fast: Learning to generate and transfer data with rectified flow},
  author={Liu, Xingchao and Gong, Chengyue and Liu, Qiang},
  journal={International Conference on Learning Representations},
  year={2022}
}

@article{lipman2022flow,
  title={Flow matching for generative modeling},
  author={Lipman, Yaron and Chen, Ricky TQ and Ben-Hamu, Heli and Nickel, Maximilian and Le, Matt},
  journal={International Conference on Learning Representations},
  year={2023}
}

@article{melnik2024video,
  title={Video diffusion models: A survey},
  author={Melnik, Andrew and Ljubljanac, Michal and Lu, Cong and Yan, Qi and Ren, Weiming and Ritter, Helge},
  journal={arXiv preprint arXiv:2405.03150},
  year={2024}
}

@inproceedings{peebles2023scalable,
  title={Scalable diffusion models with transformers},
  author={Peebles, William and Xie, Saining},
  booktitle={IEEE/CVF International Conference on Computer Vision},
  pages={4195--4205},
  year={2023}
}

@article{batifol2025flux,
  title={FLUX. 1 Kontext: Flow Matching for In-Context Image Generation and Editing in Latent Space},
  author={Batifol, Stephen and Blattmann, Andreas and Boesel, Frederic and Consul, Saksham and Diagne, Cyril and Dockhorn, Tim and English, Jack and English, Zion and Esser, Patrick and Kulal, Sumith and others},
  journal={arXiv e-prints},
  pages={arXiv--2506},
  year={2025}
}

@book{doob1984classical,
  title={Classical potential theory and its probabilistic counterpart},
  author={Doob, Joseph L and Doob, JI},
  volume={262},
  year={1984},
}

@article{lin2024fast,
  title={Fast and globally consistent normal orientation based on the winding number normal consistency},
  author={Lin, Siyou and Shi, Zuoqiang and Liu, Yebin},
  journal={ACM Transactions on Graphics },
  volume={43},
  number={6},
  pages={1--19},
  year={2024}
}

@inproceedings{zhang2025multigo,
  title={Multigo: Towards multi-level geometry learning for monocular 3d textured human reconstruction},
  author={Zhang, Gangjian and Yao, Nanjie and Zhang, Shunsi and Zhao, Hanfeng and Pang, Guoliang and Shu, Jian and Wang, Hao},
  booktitle={IEEE/CVF Conference on Computer Vision and Pattern Recognition},
  pages={338--347},
  year={2025}
}

@inproceedings{zhang2025idol,
    author    = {Zhuang, Yiyu and Lv, Jiaxi and Wen, Hao and Shuai, Qing and Zeng, Ailing and Zhu, Hao and Chen, Shifeng and Yang, Yujiu and Cao, Xun and Liu, Wei},
    title     = {IDOL: Instant Photorealistic 3D Human Creation from a Single Image},
    booktitle = {IEEE/CVF Conference on Computer Vision and Pattern Recognition},
    
    year      = {2025},
    pages     = {26308-26319}
}

@inproceedings{saito2020pifuhd,
  title={{PIFuHD}: Multi-level pixel-aligned implicit function for high-resolution 3d human digitization},
  author={Saito, Shunsuke and Simon, Tomas and Saragih, Jason and Joo, Hanbyul},
  booktitle={ IEEE/CVF Computer Vision and Pattern Recognition Conference },
  pages={84--93},
  year={2020}
}
}

\vfill

\end{document}